\pdfoutput=1
\documentclass[11pt,a4paper]{article}
\usepackage[hyperref]{emnlp2020}
\usepackage{times}
\usepackage{latexsym}
\usepackage{multirow}

\usepackage{microtype}

\aclfinalcopy 

\title{Phonemer at WNUT-2020 Task 2: Sequence Classification Using COVID Twitter BERT and Bagging Ensemble Technique based on Plurality Voting}

\author{Anshul Wadhawan \\
  Flipkart Private Limited \\
  \texttt{anshul.wadhwan@flipkart.com} } 

\date{}

\begin{document}
\maketitle
\begin{abstract}
This paper presents the approach that we employed to tackle the EMNLP WNUT-2020 Shared Task 2 : Identification of informative COVID-19 English Tweets. The task is to develop a system that automatically identifies whether an English Tweet related to the novel coronavirus (COVID-19) is informative or not. We solve the task in three stages. The first stage involves pre-processing the dataset by filtering only relevant information. This is followed by experimenting with multiple deep learning models like CNNs, RNNs and Transformer based models. In the last stage, we propose an ensemble of the best model trained on different subsets of the provided dataset. Our final approach achieved an F1-score of 0.9037 and we were ranked sixth overall with F1-score as the evaluation criteria.
\end{abstract}

\section{Introduction}
Up till mid-June 2020, the coronavirus pandemic has caused 445K deaths and has infected more than 8.2M people belonging to 215 regions and countries. This has led to a surge of panic and fear among people all around the globe. Recently, there has been rapid development in building monitoring systems (e.g. The Johns Hopkins Coronavirus Dashboard) to track any news regarding the outbreak, to let users know of any information related to the coronavirus for example, new cases emerging near the user’s location. Most of the official sources from where the information is released are not updated very frequently and the information reported by such organizations may be stale, for example, WHO updates the information regarding the virus only once a day. These monitoring systems tend to use social network data like posts from Facebook or tweets from Twitter as an alternative source of information relating to the pandemic, generally by scraping relevant information or by crowd sourcing. However, due to increased panic and emotions among people, social media is flooded with massive amounts of data, e.g. about 4M COVID-19 English tweets are posted daily on twitter. But the major issue regarding this is that the majority of such tweets are uninformative and thus, do not impart useful information that can be used. Hence, any application that requires to update the news has to first filter these tweets to detect which among the tweets are actually useful. Manual approaches for this filtering task are not only cumbersome, frustrating and ineffective for vast amounts of data, but also costly. This calls for automated systems which can filter the information given huge amounts of mixed data, and thus serve as the motivation for the shared task. Although a lot of work has been done on the lines of sequence classification on general English texts \cite{agarwal, hamid} , since the COVID-19 outbreak has grown in a short while, not many systems which deal with COVID-19 related texts have been developed.

In this paper, we describe our approach to tackle the WNUT 2020 shared task 2. The paper is structured as follows: Section 2 talks about the problem statement and provided dataset. Section 3 describes a step-by-step methodology process that we employ. Section 4 explains the experiments that were carried out along with a detailed discussion of the dataset, system settings and results of our experiments. Section 5 provides a brief conclusion of the paper along with the future scope of our research.

\section{Task Definition}

The WNUT-2020 Shared Task 2 \cite{covid19tweet} is based on a sequence classification problem wherein the aim is to identify whether an English Tweet related to the novel coronavirus (COVID-19) is informative or not. A tweet is said to be informative if it provides information regarding recovered, suspected, confirmed, death cases, location or travel history of cases.

The goals of the shared task are:
\begin{enumerate}
\item To develop a language processing task that potentially impacts research and downstream applications.
\item To provide the community with a new dataset for identifying informative COVID-19 English Tweets. 
\end{enumerate}

To achieve the goals of the shared task, a dataset of 10K COVID-19 English tweets are provided, in which 4719 tweets are labelled INFORMATIVE and 5281 tweets are labelled UNINFORMATIVE. Each tweet is annotated by three independent annotators with an inter-annotator agreement score of Fleiss' Kappa at 0.818. The 10K dataset is divided into training/validation/test sets in the ratio 70/10/20 with distribution as shown in table 1.

\begin{table}[]
\centering
\begin{tabular}{|c|c|c|c|}
\hline
               & Total & Positive & Negative \\ \hline
Training set   & 7000  & 3303     & 3697     \\ \hline
Validation set & 1000  & 472      & 528      \\ \hline
Test set       & 2000  & 944      & 1056     \\ \hline
\end{tabular}
\caption{Data Distribution}
\label{table:1}
\end{table}

Systems are evaluated using standard evaluation metrics, including accuracy, precision, recall and F1-score. However, the submissions are ranked by F1-score.

\section{Methodology}

We have split the proposed methodology in three steps- data preprocessing, deep learning models for sequence classification and ensemble process (bagging). The code corresponding to each of the steps has been made available online\footnote{\url{https://github.com/anshulwadhawan/BERT\_for\_sequence\_classification\_COVID}} to facilitate further research. 

\subsection{Data Pre-Processing}

The dataset provided in the WNUT shared task is not suitable to be processed directly by the models we plan to implement. This is due to the fact that the texts provided are tweets which are directly fetched from the website and users from all over the world have different ways of expressing their opinions. On manually going through the dataset, we find that the texts are very diverse in the sense that many users post non-ascii characters such as emoticons, slang words for informal tweets, and spelling errors in words, etc. The dataset contains URLs replaced by the tag HTTPURL and user mentions replaced by the tag @USER. Apart from this, newline characters are also present within tweets. All the above discrepancies add to noise and do not contribute to being appropriate features for sequence classification.   

In order to clean this data, we perform the following cleaning operations :
\begin{enumerate}
\item Remove all non-ascii characters i.e. characters belonging to the range [\textbackslash x00-\textbackslash x7f]. We determined this range by parsing through the dataset and recording all non-ascii characters.
\item Remove all newline (\textbackslash n) and tab (\textbackslash t) characters.
\item Remove all HTTPURL and @USER tags.

\end{enumerate}

\subsection{Deep Learning Models}

Deep learning techniques have recently shown great results in the domain of computer vision \cite{krizhevsky} and speech recognition \cite{graves}. As far as natural language processing is concerned, most of the work involving deep learning makes use of word vector representations \cite{bengio,yih2011,mikolov} to carry out finer tasks like classification.

\subsubsection{CNNs} \cite{kim} Convolutional Neural Networks are used to operate on local features with the help of convolving filters. CNNs have not only shown promising results in the domain of computer vision \cite{lecun}, but they have also been utilized extensively for NLP tasks like search query retrieval \cite{shen}, semantic parsing \cite{yih2014}, sentence modeling \cite{kalchbrenner}, and other traditional NLP tasks \cite{collobert}.
\subsubsection{RNNs}
{\bf LSTM} : LSTMs have shown great results in sequence classification problems like political sentiment classification \cite{rao}, by capturing the appropriate context. Also, they work towards solving the vanishing gradient problem \cite{hochreiter}.
{\bf BiLSTM} : Bi-directional LSTMs have tackled a variety of sequence classification tasks \cite{wang} by considering the fact that context of a word depends on the words occurring before it as well as those occurring after it.
{\bf Attention based BiLSTM} : By including attention to a BiLSTM, we try to find out the specific words which have the greatest impact to the overall sentiment of the sequence under consideration. 
\subsubsection{Transformer based models}
{\bf BERT (bert-base-cased)}: \cite{bert} BERT is a bidirectional transformer based model pre-trained on a huge corpus of Wikipedia and Toronto Book Corpus which uses a combination of objectives meant for the tasks of next sentence prediction and masked language modeling.
{\bf RoBERTa (roberta-base)}: \cite{roberta} It is built on top of BERT by removing the next sentence prediction objective, changing key hyperparameters and training with increased learning rate values and batch sizes.
{\bf ALBERT (albert-base-v2)}: \cite{albert} This is another variation of BERT which tries to increase the training speed of BERT and lower memory utilization by repeating layers which are split among groups and splitting the embedding matrix into two.
{\bf XLNet (xlnet-base-cased)}: \cite{xlnet} This model extends over the Transformer-XL model by learning bidirectional contexts and maximizing the likelihood over different permutations of the input sequence factorization order after pre-training.
{\bf XLM (xlm-mlm-en-2048)}: \cite{xlm}  This is a transformer based model with an option to choose the objective functions from the tasks of masked language modeling, casual language modeling, and translation language modeling.
{\bf COVID Twitter BERT (ct-bert)}: \cite{ctbert} COVID-Twitter-BERT (CT-BERT) is a transformer-based model pre-trained on a corpus of 22.5M (633M tokens) COVID-19 related tweets.

\subsection{Ensemble Process - Bagging}

We merge the training and validation datasets provided in the task to create a global dataset. Then, we shuffle this global dataset and split it into training and validation datasets with the same ratio. This process is repeated seven times to create seven sets of training and validation datasets, each of which have a random class distribution. The best performing model, based on validation scores on default training and validation split provided in the task, is trained from scratch on each of these seven sets of training and validation splits, and the predictions are recorded. Once we have the seven sets of predictions, we use a max-voting algorithm that is based on calculating mode of the seven predictions for each test instance to produce the final predictions.   

\section{Experiments}

We experiment with ten deep learning models with the provided training and validation splits. Based on the scores produced above, we evaluate the final test predictions by training the best performing model with the seven synthesised randomly shuffled versions of the dataset, followed by merging the output predictions made on the test set.
In this section, we present the dataset distribution, experimental settings, evaluation metrics, results and a brief analysis of the proposed system. 

\subsection{Dataset}

The class-wise distribution in the training and validation splits of the provided as well as the shuffled datasets are shown in Table 2.

\begin{table}[]
\centering
\begin{tabular}{|c|c|c|c|c|}
\hline
\multicolumn{1}{|l|}{\multirow{2}{*}{}} & \multicolumn{2}{c|}{Train Set} & \multicolumn{2}{c|}{Val Set} \\ \cline{2-5} 
\multicolumn{1}{|l|}{} & Pos & Neg & Pos & Neg \\ \hline
Shuffle1 & 3285 & 3715 & 490 & 510 \\ \hline
Shuffle2 & 3294 & 3706 & 481 & 519 \\ \hline
Shuffle3 & 3305 & 3695 & 470 & 530 \\ \hline
Shuffle4 & 3293 & 3707 & 482 & 518 \\ \hline
Shuffle5 & 3313 & 3687 & 462 & 538 \\ \hline
Shuffle6 & 3299 & 3701 & 476 & 524 \\ \hline
Shuffle7 & 3293 & 3707 & 482 & 518 \\ \hline
\end{tabular}
\caption{Data Distribution}
\label{tab:my-table}
\end{table}

\subsection{System Settings}

For training the CNN, LSTM and BiLSTMs, word vectors for english language pre-trained on Common Crawl\footnote{\url{https://commoncrawl.org/}} and Wikipedia\footnote{\url{https://www.wikipedia.org/}} are downloaded\footnote{\url{https://dl.fbaipublicfiles.com/fasttext/vectors-crawl/cc.en.300.bin.gz}} and used using FastText\footnote{\url{https://fasttext.cc/docs/en/crawl-vectors.html}} library. These word vectors are used to create the embedding matrix which is further used for transforming the words of the input sentence. We use binary cross entropy loss function and adam optimizer for all the CNN and RNN models. All the layers except the last layer have relu activation function. Since the problem is a binary classification, we use sigmoid activation function in the last layer. The training session is run for a total of 20 epochs and early stopping was inculcated in case of successive unproductive(in terms of f1-score) iterations. For all proposed RNNs, dropout of 0.2 and number\_of\_units of 150 are found to be the most effective. 
To fine-tune the transformer based models, we use pre-trained models, namely bert-base-cased, roberta-base, albert-base-v2, xlnet-base-cased and xlm-mlm-en-2048. We use hugging-face\footnote{\url{https://huggingface.co/transformers/}} API to train all the transformer based models. We use a learning\_rate of 4e-5, epsilon\_parameter\_for\_adam\_optimizer of 1e-8, maximum\_sequence\_length of 128 and a batch\_size of 8 due to hardware limitations. We train the models for 10 epochs and evaluate the model’s performance on the validation set after every epoch.

\subsection{Results and Analysis}

The performance results of the proposed models on the given validation dataset in terms of f1-score(F1), precision(P), recall(R) and accuracy(A) have been presented in table 3.

\begin{table}[]
\centering
\begin{tabular}{|l|l|l|l|l|}
\hline
           & \multicolumn{1}{c|}{\color[HTML]{000000}F1} & \multicolumn{1}{c|}{\color[HTML]{000000}P} & \multicolumn{1}{c|}{\color[HTML]{000000}R} & \multicolumn{1}{c|}{\color[HTML]{000000}A} \\ \hline
\color[HTML]{000000} 
CNN        & 0.787                                        & 0.805                                       & 0.771                                       & 0.804                                       \\ \hline
\color[HTML]{000000}
LSTM       & 0.807                                        & 0.850                                       & 0.769                                       & 0.827                                       \\ \hline
\color[HTML]{000000}
BiLSTM     & 0.822                                        & 0.806                                       & 0.838                                       & 0.829                                       \\ \hline
\color[HTML]{000000}
Att-BiLSTM & 0.823                                        & 0.773                                       & 0.881                                       & 0.822                                       \\ \hline
\color[HTML]{000000}
BERT       & 0.891                                        & 0.875                                       & 0.908                                       & 0.896                                       \\ \hline
\color[HTML]{000000}
RoBERTa    & 0.899                                        & 0.886                                       & 0.913                                       & 0.904                                       \\ \hline
\color[HTML]{000000}
ALBERT     & 0.844                                        & 0.851                                       & 0.836                                       & 0.854                                       \\ \hline
\color[HTML]{000000}
XLNet      & 0.892                                        & 0.864                                       & 0.921                                       & 0.895                                       \\ \hline
\color[HTML]{000000}
XLM        & 0.870                                        & 0.850                                       & 0.891                                       & 0.875                                       \\ \hline
\color[HTML]{000000}
CT-BERT    & \textbf{0.914}                               & 0.869                                       & 0.963                                       & \textbf{0.915}                              \\ \hline
\end{tabular}
\caption{Model Scores on Validation Set}
\label{table:3}
\end{table}

We can conclude the following from table 3:
\begin{enumerate}
\item RNN based models perform better than CNN due to their context capturing potential.
\item Transformer based models perform better than both CNN and RNN based models due to more parallelization because of the fact that they don’t need to traverse the input in order.
\item The COVID Twitter BERT (CT-BERT) outperforms all the other models by a significant margin. Higher recall is one major observation in this case. This is evident from the fact that the model is pre-trained on 22.5M COVID-19 related tweets.

\end{enumerate}

From the above results, we choose CT-BERT to be the primary model to train on the seven randomly shuffled datasets as well as carry out inferences on the unseen dataset. The ensemble of produced predictions results in an F1-score of 0.9037 on the test dataset. The best performing model i.e. CT-BERT on being trained over the global dataset results in an F1-score of 0.8954 on the test dataset. This shows that by inculcating the bagging technique, a boost of 0.83\% in F1-score is seen. This can be credited to the fact that by creating seven sets of randomly selected training examples, we essentially cover the entire global dataset. Although we exclude portions of the dataset by excluding randomly selected instances from the training dataset, we cover all the instances in the global dataset by selecting those labels which are predicted by majority of the seven trained CT-BERT models. This can be deduced from the assumption that a particular instance which is absent in the training set of one of the seven models is likely to be present in the training set of most of the remaining models.

\section{Conclusion and Future Work}

In this paper, we provide a detailed description of our approach to solve the EMNLP WNUT-2020 Shared Task 2. Our approach involves processing in three stages. In the first stage, we pre-process the provided dataset by cleaning and extracting only relevant information from the provided text. In the second stage, we experiment with several deep neural networks like CNN, RNNs and Transformer based networks like XLNet, XLM, BERT and its different variations. In the final phase, we introduce the idea of ensemble learning (bagging) to our solution which is a major improvement over the individual model. We submitted an ensemble and an individual system based on the CT-BERT model as our final entries to the shared task. The ensemble approach fetches us a private leaderboard rank of 6 with F1-score as the evaluation criteria. Our system promotes the usage of transformer based models pre-trained on relevant corpora and ensemble learning with as many candidate models as feasible. In future, we aim to explore the usage of non-ascii characters like emoticons as features for classification and imparting ensemble learning through an end-to-end deep learning solution.

\bibliographystyle{acl_natbib}
\bibliography{emnlp2020}
\end{document}